\documentclass[runningheads]{llncs}

\usepackage{graphicx,amsmath, amssymb,multirow, psfrag}
\begin{document}

\title{LSTM-Based Anomaly Detection: \\ Detection Rules from Extreme Value Theory}
\titlerunning{LSTM-Based Anomaly Detection}
\author{Neema Davis \and
Gaurav Raina \and
Krishna Jagannathan}
\authorrunning{N. Davis et al.}
\institute{Department
of Electrical Engineering, Indian Institute of Technology Madras, Chennai 600 036, India \\
\email{\{ee14d212, gaurav, krishnaj\}@ee.iitm.ac.in}}
\maketitle            
\begin{abstract}
In this paper, we explore various statistical techniques for anomaly detection in conjunction with the popular Long Short-Term Memory (LSTM) deep learning model for transportation networks. We obtain the prediction errors from an LSTM model, and then apply three statistical models based on (i) the Gaussian distribution, (ii) Extreme Value Theory (EVT), and (iii) the Tukey's method. Using statistical tests and numerical studies, we find strong evidence against the widely employed Gaussian distribution based detection rule on the prediction errors. Next, motivated by fundamental results from Extreme Value Theory, we propose a detection technique that does not assume any parent distribution on the prediction errors. Through numerical experiments conducted on several real-world traffic data sets, we show that the EVT-based detection rule is superior to other detection rules, and is supported by statistical evidence.

\end{abstract}
\section{Introduction}

 Mobility modeling can aid the design of sustainable transportation systems, making it a crucial part of Intelligent Transportation Systems (ITS). A popular example can be the Mobility-On-Demand service providers such as e-hailing taxis that rely on efficient passenger mobility modeling for rerouting their drivers. The demand for taxis changes dynamically with daily human mobility patterns, along with other non-periodic events. 
While short-term taxi demand forecasting models may learn periodic patterns in demand \cite{davis2018ataxi}, \cite{liao2018large}, they are normally unable to accurately capture non-periodic mobility events. It is necessary to detect these unusual events as they often indicate useful, and critical information that can yield instructive insights, and help to develop more accurate prediction models and strategies. This task of finding patterns in data that do not conform to a certain expected behavior is called \emph{anomaly detection}. Transportation networks present several situations where one may find anomalous behavior and patterns. For example, sudden spikes in taxi demand might indicate the ending of a concert, drops in traffic speed might be the effect of an unprecedented event such as a road accident, and so on. 

The identification of anomalies has been traditionally tackled using statistical and machine learning techniques; for a survey, see \cite{chandola2009anomaly}. The substantial advances made by deep learning methods in recent years, in various machine learning problems, have encouraged researchers to explore them for anomaly detection as well \cite{chalapathy2019deep}. The suitability of deep learning models for anomaly detection stems from their unsupervised learning nature and the ability to learn highly complex non-linear sequences. Popular deep learning models such as the Long Short-Term Memory (LSTM) \cite{malhotra2015long} and related Recurrent Neural Networks (RNNs) \cite{lu2017traffic} have shown superior anomaly detection performance compared to traditional anomaly detection models \cite{cheng2018multi}.  When presented with normal non-anomalous data, the LSTM can learn and capture the normal behavior of the system. Later, when the LSTM encounters a data instance that deviates significantly from the rest of the set, it generates a high prediction error, suggesting anomalous behavior. This form of prediction-based anomaly detection has found applications in cardiology \cite{chauhan2015anomaly}, automobiles \cite{taylor2016anomaly}, radio communications \cite{o2016recurrent}, Cyber-Physical Systems (CPS) \cite{inoue2017anomaly}, and telemetry \cite{hundman2018detecting}, among others.

Prediction-based anomaly detection requires the application of a set of detection rules on the prediction errors. Usually, this is performed by employing a traditional anomaly detection algorithm on top of the prediction errors. Often, the detection rule involves assuming an underlying parametric distribution on the prediction errors, which is mostly Gaussian \cite{malhotra2015long}. In addition to being computationally efficient and mathematically tractable, a distribution based detection rule does not require large memory storage (unlike clustering-based approaches) or suitable kernel functions (unlike Support Vector Machines). If the assumptions regarding the underlying data distribution hold, this technique provides a statistically justifiable solution for anomaly detection. However, its disadvantage is also rooted in this assumption; faithful detection is possible only if the data conforms to a particular distribution.

Given the rise in prediction-based anomaly detection methods and related research \cite{chalapathy2019deep}, it is essential to place increased emphasis on post-prediction error evaluation methods that have received comparatively less focus yet are instrumental for accurate anomaly detection. Since the prediction errors are often assumed to be Gaussian \cite{malhotra2015long}, we investigate this assumption by conducting statistical tests on seven diverse real-world data sets. In particular, we compare this Gaussian-based decision rule against decision rules based on Extreme Value Theory (EVT) \cite{beirlant2006statistics}. An important result from EVT suggests that the extreme values of any distribution follow a Generalized Pareto Distribution (GPD), regardless of the parent distribution. This result allows us to develop a parametric distribution based detection rule on the prediction errors, without making critical assumptions about the input data distribution. Currently, EVT-based detection rules have not been explored for LSTM-based anomaly detection, setting our work apart from the existing literature.  Further, we compare these two distribution-based detection rules against a detection rule based on Tukey's method \cite{tukeyw} that makes no explicit assumptions on the distribution.



\subsection{Our Contributions}
We compare three detection rules for LSTM-based anomaly detection: (i) the Gaussian-based detection rule that makes assumptions about the parent distribution, (ii) the EVT-based detection rule that does not assume a parent distribution but makes assumptions about the distribution of the tail, and (iii) the Tukey's method based detection rule that does not make any assumptions on the distribution. 
The major findings of this paper are as follows:
\begin{enumerate}
    \item Using the Shapiro-Wilk test for Normality, we reject the null hypothesis that the prediction errors follow the Normal distribution with high confidence.
    \item We find that the EVT-based detection rule has superior anomaly detection performance compared to the Gaussian-based detection rule and Tukey's method based detection rule.
    \item The false-positive regulator of the EVT-based detection rule has a lower variance than that of the Gaussian-based detection rule, resulting in faster parameter tuning.
\end{enumerate}
The rest of the paper is organized as follows. Section \ref{data} provides a brief description of the data sets. Our anomaly detection methodology, along with the three detection rules and evaluation metrics are explained in Section \ref{admodels}. This is followed by a discussion on the statistical tests in Section \ref{statistical}. The experimental settings and results are elaborated in Section \ref{experiments}. We summarize our contributions in Section \ref{conclusions}.

\section{Data Sets} \label{data}
We consider seven real-world data sets for our comparison study: two taxi demand data sets, three traffic-based data sets, and two data sets from other application domains. The travel time, occupancy and speed data sets are real-time data, obtained from a traffic detector and collected by the Minnesota Department of Transportation. Discussions on the traffic data sets and the temperature sensor data set are available at the Numenta Anomaly Benchmark GitHub repository\footnote{\label{note1}https://github.com/numenta/NAB/tree/master/data}. Brief descriptions of all data sets used are given below. 
\begin{enumerate}
    \item Vehicular Travel Time: The data set is obtained from a traffic sensor and has 2500 readings from July 10, 2015, to September 17, 2015, with eight marked anomalies.  
    \item Vehicular Speed: The data set contains the average speed of all vehicles passing through the traffic detector. A total of 1128 readings for the period September 8, 2015 - September 17, 2015, is available. There are three marked unusual sub sequences in the data set. 
    \item Vehicular Occupancy: There are a total of 2382 readings indicating the percentage of the time, during a 30-second period, that the detector sensed a vehicle. The data is available for a period of 17 days, from September 1, 2015, to September 17, 2015, and has two marked anomalies. 
    \item New York City Taxi Demand \cite{nyc}: The publicly available data set contains the pick-up locations and time stamps of street hailing yellow taxi services from the period of January 1, 2016, to February 29, 2016. We pick three time-sequences with clearly evident anomalies from data aggregated over 15 minute time periods in 1 $\text{km}^2$ grids. 
    \item Bengaluru Taxi Demand: This data set is obtained from a private taxi service provider in Bengaluru, India and has GPS traces of passengers booking a taxi by logging into the service provider's mobile application. Similar to the New York City data set, this data is also available for January and February 2016. We aggregate the data over 15 minute periods in 1 $\text{km}^2$ grids and pick three sequences with clearly visible anomalies.
    \item Electrocardiogram (ECG) \cite{keogh2005hot}: There are a total of 18000 readings, with three unusual sub sequences labeled as anomalies. The data set has a repeating pattern, with some variability in the period length.
    \item Machine Temperature: This data set contains temperature sensor readings from an internal component of a large industrial machine. The readings are for the period between December 2, 2013, to February 19, 2014. There are a total of 22695 readings taken every 5 minutes, consisting of four anomalies with known causes.
\end{enumerate}

\section{Anomaly Detection} \label{admodels}
This section outlines our prediction-based anomaly detection, along with the detection strategies.

\subsection{Prediction Model}
We use the Long Short-Term Memory (LSTM) network \cite{gers2000learning} as the time-series prediction model. The most recent $l_b$ values of every data set are fed into the model to output $l_a$ future values. We refer to parameters $l_b$, $l_a$ as look-back and look-ahead respectively. Dropout and early stopping are employed to avoid over-fitting. Before training any neural network model, it is necessary to set suitable values for various hyper-parameters. These parameters define the high-level features of the model, such as its complexity, or capacity to learn. For example, in a neural network model, the important hyper-parameters are the number of hidden recurrent layers, the dropout values, the learning rate, and the number of units in each layer. We use the Tree-structured Parzen Estimator (TPE) Bayesian Optimization \cite{bergstra2011algorithms} to select these hyper-parameters. The output layer is a fully connected dense layer with linear activation. The Adam optimizer is used to minimize the Mean Squared Error.

Each data set is divided into a training set, a hold-out validation set, and a test set. The training set is assumed to be free of anomalies. This is a reasonable assumption in real-world anomaly detection scenarios where the occurrence of anomalous behavior is rare compared to the occurrence of instances of normal behavior. The validation and test set are mixtures of anomalous and non-anomalous data instances. The prediction model is trained on normal data without any anomalies, i.e., on the training data, so that it learns the normal behavior of the time-series. Once the model is trained, anomaly detection is performed by using the prediction errors as anomaly indicators. In this study, the prediction error is defined as the absolute difference between the input received at time $t$ and its corresponding prediction from the model. Next, we discuss three techniques (detection rules) by which the prediction errors can be used to set an anomaly threshold. If any prediction error value lies outside of the threshold, then the corresponding input value can be considered as a possible anomaly.

\subsection{Gaussian-based Detection}
The prediction errors from the training data are assumed to follow a Gaussian distribution. The mean, $\mu$, and variance, $\sigma^2$, of the Gaussian distribution are computed using the Maximum Likelihood Estimation (MLE). The Log Probability Densities (Log PDs) of errors are calculated based on these estimated parameters and used as anomaly scores \cite{singh2017anomaly}. A low value of Log PD indicates that the likelihood of an observation being an anomaly is high. A validation set containing both normal data and anomalies is used to set a threshold $\tau_{g}$ on the Log PD values. The threshold is chosen such that it can separate all the anomalies from normal observations while incurring as few false positives as possible. The threshold is then evaluated on a separate test set.

\subsection{EVT-based Detection}
Let $X$ be a random variable and $F(x) = P(X \leq x)$ be its Cumulative Distribution Function (CDF). The tail of the distribution is given by $\Tilde{F}(x) = P(X > x)$. 
A key result from the Extreme Value Theory (EVT) \cite{beirlant2006statistics} shows that the distribution of the extreme values is not highly sensitive to the parent data distribution. This enables us to accurately compute probabilities without first estimating the underlying distribution. Under a weak condition, the extreme events have the same kind of distribution, regardless of the original one, known as the Extreme Value Distribution (EVD):
\begin{equation}
    G_\gamma : y \rightarrow \text{exp} \Big ( -\Big(1+\gamma y\Big)^{-\frac{1}{\gamma}} \Big),\ \gamma \in \mathbb{R},\ 1 + \gamma y > 0,
\end{equation}
where $\gamma$ is the extreme value index of the distribution. By fitting an EVD to the unknown input distribution tail, it is then possible to evaluate the probability of potential extreme events. In some recent work \cite{siffer2017anomaly}, the authors use results from EVT to detect anomalies in a uni-variate data stream, following the Peaks-Over-Threshold (POTs) approach. Based on an initial threshold $t$, the POTs approach attempts to fit a Generalized Pareto Distribution (GPD) to the excesses, $X-t$. Once the parameters of the GPD are obtained using MLE, the threshold can be computed as:
\begin{equation}
    \tau_e = t+ \frac{\hat{\sigma}}{\hat{\gamma}} \Big( \Big(\frac{qn}{N_t} \Big)^{-\hat{\gamma}} -1 \Big ),
\end{equation}
where $\hat{\gamma}$ and $\hat{\sigma}$ are the estimated parameters of the GPD, $q$ is some desired probability, $n$ is the total number of observations, $N_t$ is the number of peaks, i.e., the number of $X_i$ s.t. $X_i > t$. We calculate $P(X > \tau_e)$ for all the observations and those data instances with $P(X > \tau_e) < q$ can be considered as plausible anomalies. We apply this methodology to the prediction errors obtained from the LSTM. The authors in \cite{siffer2017anomaly} recommend choosing $q$ in the range [$10^{-3}$, $10^{-5}$] and initial $t$ as the 98\% quantile, which we follow in our study. 

\subsection{Tukey's Method Based Detection}
Tukey's method \cite{tukeyw} uses quartiles to define an anomaly threshold. It makes no distributional assumptions and does not depend on a mean or a standard deviation. In Tukey's method, a possible outlier lies outside the threshold $\tau_{t} = Q_3 + 3 \times (Q_3 - Q_1)$, where $Q_1$ is the lower quartile or the  $25^{th}$ percentile, and $Q_3$ is the upper quartile or the $75^{th}$ percentile. The prediction errors from the training, validation and test sets are concatenated, and the lower and upper quartiles are calculated. The values lying outside $\tau_{t}$ are identified as possible outliers.

\subsection{Evaluation Metrics}
We consider three evaluation metrics for comparing the detection rules: (i) Precision, $P$, which is the ratio of true positives to the sum of true positives and false positives, (ii) Recall, $R$, which is the ratio of true positives to the sum of true positives and false negatives, and (iii) F1-score, $F1$, which is the harmonic mean of Precision and Recall. Since F1-score summarizes both Precision and Recall, we consider the detection rule with the highest $F1$ as the superior anomaly detection technique. 
True positives refer to the correctly predicted anomalies. False positives are the non-anomalies that we incorrectly identify as being anomalies. False negatives refer to the anomalies incorrectly identified as non-anomalous instances. 

\section{Statistical Tests} \label{statistical}
We conduct two sets of statistical tests: (i) the Shapiro-Wilk test \cite{shapiro1965analysis} for testing the Normality of the prediction errors, and (ii) the Anderson-Darling test \cite{stephens1974edf} for checking the compliance of the tail distribution to a Generalized Pareto Distribution (GPD). 

The Shapiro-Wilk test \cite{shapiro1965analysis} calculates a $W$ statistic that tests whether a sample comes from a Normal distribution. The $W$ statistic measures the correlation between the given data and ideal normal scores. 
If the p-value is less than the chosen significance level (typically less than 0.05), then the null hypothesis can be rejected and there is evidence that the data tested are not Normally distributed. The Anderson-Darling test \cite{stephens1974edf} is used to assess whether a sample of the data comes from a specific probability distribution. 
The test statistic $A^2$ measures the distance between the hypothesized distribution and the empirical CDF of the data. Based on the test static and the p-values obtained, the null hypothesis that the data follow a specified distribution can (cannot) be rejected. The Anderson-Darling is a modification of the Kolmogorov-Smirnov (K-S) test \cite{massey1951kolmogorov} and gives more weight to the tails than does the K-S test. 

The p-values obtained by conducting the Shapiro-Wilk test on the prediction errors and the Anderson-Darling test on the excesses $X-t$ are given in Table \ref{pval}. The null hypothesis is rejected for p-values less than 0.001. For all the data sets under study, based on the p-values of the Shapiro-Wilk test, we rejected the null hypothesis that the prediction errors follow a Gaussian distribution. At the same time, statistical evidence from the Anderson-Darling test suggests that the tail distributions of the various prediction errors tend to follow GPD. Hence, while assuming a Normal distribution on the prediction errors may not be suitable for LSTM-based hybrid anomaly detection, a GPD seems to be a more reasonable fit. 
\begin{table}[]
\caption{P-values obtained from the statistical tests. The decision to reject the null hypothesis is taken when the p-values lie below 0.001. The null hypothesis that the prediction errors follow a Gaussian distribution is rejected, and that the tails of the prediction errors follow a Generalized Pareto Distribution is accepted.}
\label{pval}
\begin{center}
{\renewcommand{\arraystretch}{1.2}
\begin{tabular}{|l||c||c|}
\hline
\multicolumn{1}{|c||}{\multirow{2}{*}{\textbf{Data Sets}}} & \multicolumn{2}{c|}{\textbf{P-values}}             \\ \cline{2-3} 
\multicolumn{1}{|c||}{}                          & Shapiro-Wilk Test & Anderson-Darling Test \\ \hline \hline
Vehicular Travel Time                           &  0.000             &   0.005                    \\ \hline
Vehicular Speed                                 &  2.38e-22         &   0.005                    \\ \hline
Vehicular Occupancy                             &  6.64e-23         &   0.37                \\ \hline
NYC Taxi Demand                                 &  2.62e-42         &   0.14                    \\ \hline
Bengaluru Taxi Demand                           &  4.45e-43         &   0.57                    \\ \hline
Electrocardiogram                                             &  0.000             &   0.002                    \\ \hline
Machine Temperature                             &  0.000             &   0.002                    \\ \hline
\end{tabular}}
\end{center}
\end{table}

\section{Experiments} \label{experiments}
As mentioned in Section \ref{admodels}, hyper-parameter optimization is performed prior to the model training process. The chosen set of parameters for each data set is given in Table \ref{settings}. We follow the same model settings as \cite{singh2017anomaly} for the ECG and Machine Temperature data sets. For the traffic speed, travel time and vehicular occupancy data sets, the limited availability of readings suggested look-back and look-ahead times of 1 each. We have over 10 million points for the New York and Bengaluru cities, allowing for a large look-back time. The considerable amount of data in these two cases allows the LSTM to learn better representations of the input data, aiding the anomaly detection process. The models ran for 100 epochs with a batch size of 64, minimizing the Mean Squared Error. 

\begin{table}[]
\caption{The experimental settings for the data sets considered. The optimal set of hyper-parameters for each data set is chosen after running the Tree-structured Parzen Estimator (TPE) Bayesian Optimization.  }
\label{settings}
\centering
{\renewcommand{\arraystretch}{1.1}
\begin{tabular}{|l||l||c|}
\hline
\multicolumn{1}{|c||}{\textbf{Data Sets}} & \multicolumn{1}{c||}{\textbf{LSTM Architecture}}                                                                                                                    & \textbf{$l_b$, $l_a$}     \\ \hline \hline
Vehicular Travel Time       & \begin{tabular}[l]{@{}l@{}}1 Recurrent layer: \{20\}, Dropout: 0.2,\\ 1 Dense layer: \{1\}, Learning rate: 0.01\end{tabular}         & 1, 1      \\ \hline
Vehicular Speed                & \begin{tabular}[l]{@{}l@{}}1 Recurrent layer: \{60\}, Dropout: 0.19,\\ 1 Dense layer: \{1\}, Learning rate: 0.0001\end{tabular}      & 1, 1     \\ \hline
Vehicular Occupancy         & \begin{tabular}[l]{@{}l@{}}1 Recurrent layer: \{50\}, Dropout: 0.23,\\ 1 Dense layer: \{1\}, Learning rate: 0.0001\end{tabular}      & 1, 1     \\ \hline
NYC Taxi Demand                & \begin{tabular}[l]{@{}l@{}}2 Recurrent layers: \{50, 20\}, Dropout: 0.4,\\ 1 Dense layer:\{24\}, Learning rate: 0.0001\end{tabular}  & 5760, 24 \\ \hline
Bengaluru Taxi Demand          & \begin{tabular}[l]{@{}l@{}}2 Recurrent layers: \{20, 10\}, Dropout: 0.25,\\ 1 Dense layer:\{24\}, Learning rate: 0.0001\end{tabular} & 5760, 24 \\ \hline
Electrocardiogram                        & \begin{tabular}[l]{@{}l@{}}2 Recurrent layers: \{60, 30\}, Dropout: 0.1,\\ 1 Dense layer:\{5\}, Learning rate: 0.05\end{tabular}      & 8, 5     \\ \hline
Machine Temperature            & \begin{tabular}[l]{@{}l@{}}2 Recurrent layers: \{80, 20\}, Dropout: 0.1,\\ 1 Dense layer: \{12\}, Learning rate: 0.1\end{tabular}     & 24, 12   \\ \hline
\end{tabular}}
\end{table}

\begin{figure*}
\centering
 \psfrag{o}{\scalebox{3.5}{Actual Readings}}
\psfrag{p}{\scalebox{3.5}{Predictions}}
\psfrag{e}{\scalebox{3.5}{Prediction Errors}}
\psfrag{t}{\raisebox{-0.4cm}{\hspace{-0.8cm}\scalebox{3.5}{Time}}}
\psfrag{m}{\raisebox{0.4cm}{\hspace{-1.5cm}\scalebox{3.5}{Magnitude}}}
\psfrag{1.0}{\scalebox{3.5}{1}}
\psfrag{879.5}{\scalebox{3.5}{880}}
\psfrag{1759.0}{\scalebox{3.5}{1760}}
\psfrag{0}{\scalebox{3.5}{0}}
\psfrag{30}{\scalebox{3.5}{30}}
\psfrag{60}{\scalebox{3.5}{60}}
        \includegraphics[angle = -90, scale = 0.21]{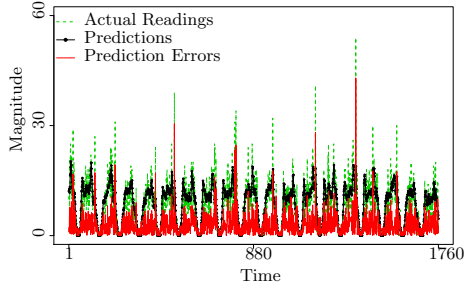}
        \caption{Performance of the LSTM model. The actual taxi demand values for one of the time-sequences from the New York City is plotted, along with the predictions and prediction errors. True anomaly is at $t$ = 1366.}
        \label{fig1}
 \end{figure*}
 
\begin{figure*}[t]
 \centering
\psfrag{l}{\scalebox{3.5}{Log PD}}
\psfrag{h}{\scalebox{3.5}{Static Threshold}}
\psfrag{d}{\scalebox{3.5}{Anomaly}}
\psfrag{t}{\raisebox{-0.4cm}{\hspace{-0.8cm}\scalebox{3.5}{Time}}}
\psfrag{ll}{\raisebox{0.4cm}{\hspace{-1.5cm}\scalebox{3.5}{Log PD}}}
\psfrag{1.0}{\scalebox{3.5}{1}}
\psfrag{879.5}{\scalebox{3.5}{880}}
\psfrag{1759.0}{\scalebox{3.5}{1760}}
\psfrag{-72}{\scalebox{3.5}{-70}}
\psfrag{-37}{\scalebox{3.5}{-35}}
\psfrag{-3}{\scalebox{3.5}{-3}}
        \includegraphics[angle = -90, scale = 0.21]{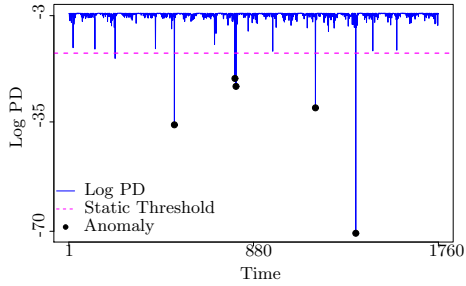}
        \caption{Performance of the Gaussian-based detection rule on prediction errors obtained from Fig \ref{fig1}. The threshold is chosen from the validation errors and is seen to result in many false-positives along with the true anomaly at $t$ = 1366. }
        \label{fig2}
 \end{figure*}
 \begin{figure*}[h!]
 \centering
  \psfrag{ee}{\scalebox{3.5}{Prediction Error}}
\psfrag{hh}{\scalebox{3.5}{Dynamic Threshold}}
\psfrag{d}{\scalebox{3.5}{Anomaly}}
\psfrag{t}{\raisebox{-0.4cm}{\hspace{-0.8cm}\scalebox{3.5}{Time}}}
\psfrag{mm}{\raisebox{0.4cm}{\hspace{-1.5cm}\scalebox{3.5}{Magnitude}}}
\psfrag{1.0}{\scalebox{3.5}{1}}
\psfrag{879.5}{\scalebox{3.5}{880}}
\psfrag{1759.0}{\scalebox{3.5}{1760}}
\psfrag{0}{\scalebox{3.5}{0}}
\psfrag{25}{\scalebox{3.5}{25}}
\psfrag{50}{\scalebox{3.5}{50}}
        \includegraphics[angle = -90, scale = 0.21]{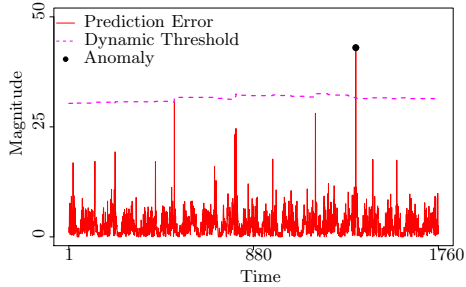}
       \caption{Performance of the EVT-based detection rule on prediction errors obtained from Fig \ref{fig1}. The detection rule chooses a threshold, without manual tuning, based on a desired probability. True anomaly detected at $t$ = 1366, without incurring any false-positives.}
       \label{fig3}
 \end{figure*}

\begin{table}[th!]
\caption{Evaluation of the detection rules on the data sets considered. The best technique for each data set is shown in bold and shows the superior performance of the EVT-based detection rule. }
\label{results}
\setlength{\tabcolsep}{2.5pt}
\centering
\scalebox{0.95}{
{\renewcommand{\arraystretch}{1.5}
\begin{tabular}{|c|c||c|c|c|c||c|c|c|c||c|c|c|}
\hline
\multicolumn{2}{|c||}{\multirow{2}{*}{\textbf{Data Sets}}}                                        & \multicolumn{4}{c||}{\begin{tabular}[c]{@{}c@{}}\textbf{Normality} \\ \textbf{Assumption}\end{tabular}} & \multicolumn{4}{c||}{\begin{tabular}[c]{@{}c@{}}\textbf{Tail Assumption}\\ \textbf{from EVT}\end{tabular}} & \multicolumn{3}{c|}{\begin{tabular}[c]{@{}c@{}}\textbf{Tukey's} \\ \textbf{Method}\end{tabular}} \\ \cline{3-13} 
\multicolumn{2}{|c||}{}                                                                 & P                   & R                  & F1                  & $\tau$              & P                   & R                  & F1                 & $q$                     & P                         & R                       & F1                       \\ \hline \hline
\multicolumn{2}{|c||}{Vehicular Travel Time}                                                      & 0.14                & 0.40               & 0.21                & -20                 & 0.33                & 0.40               & \textbf{0.36}               & $10^{-4}$               & 0.04                      & 0.60                    & 0.07                     \\ \hline
\multicolumn{2}{|c||}{Vehicular Speed}                                                  & 0.58                & 1.0                & 0.73                & -18                 & 0.75                & 0.85               & \textbf{0.79}               & $10^{-3}$               & 0.75                      & 0.85                    & \textbf{0.79}                     \\ \hline
\multicolumn{2}{|c||}{Vehicular Occupancy}                                                        & 1.0                 & 1.0                & 1.0                 & -23                 & 1.0                 & 1.0                & \textbf{1.0}                & $10^{-5}$               & 0.33                      & \textbf{1.0}                     & 0.5                      \\ \hline
\multirow{3}{*}{\begin{tabular}[c]{@{}c@{}}NYC Taxi \\ Demand\end{tabular}}       & T1 & 1.0                 & 1.0                & \textbf{1.0}                 & -19                 & 1.0                 & 1.0                & \textbf{1.0}                & $10^{-5}$               & 1.0                       & 0.14                    & 0.25                     \\ \cline{2-13} 
                                                                                  & T2 & 0.2                 & 1.0                & 0.33                & -17                 & 1.0                 & 1.0                & \textbf{1.0}                & $10^{-5}$               & 0.07                      & 1.0                     & 0.14                     \\ \cline{2-13} 
                                                                                  & T3 & 0.75                & 1.0                & \textbf{0.85}                & -15                 & 0.75                & 1.0                & \textbf{0.85}               & $10^{-5}$               & 0.5                       & 1.0                     & 0.66                     \\ \hline
\multirow{3}{*}{\begin{tabular}[c]{@{}c@{}}Bengaluru \\ Taxi Demand\end{tabular}} & T1 & 1.0                 & 0.4                & 0.57                & -25                 & 1.0                 & 1.0                & \textbf{1.0}                & $10^{-4}$               & 0.31                      & 1.0                     & 0.47                     \\ \cline{2-13} 
                                                                                  & T2 & 0.33                & 1.0                & \textbf{0.5}                 & -18                 & 0.33                & 1.0                & \textbf{0.5}                & $10^{-4}$               & 0.04                      & 1.0                     & 0.07                     \\ \cline{2-13} 
                                                                                  & T3 & 0.6                 & 0.5                & 0.54                & -25                 & 0.57                & 0.66               & \textbf{0.61}               & $10^{-4}$               & 0.15                      & 0.83                    & 0.26                     \\ \hline
\multicolumn{2}{|c||}{Electrocardiogram}                                                              & 0.50                & 0.23               & 0.32                & -23                 & 0.50                & 0.28               & {0.36}               & $10^{-4}$               & 0.42                      & 0.57                    & \textbf{0.49}                     \\ \hline
\multicolumn{2}{|c||}{Machine Temperature}                                                    & 0.004               & 0.50               & 0.009               & -19                 & 0.10                & 0.50               & \textbf{0.16}               & $10^{-4}$               & 0.002                     & 0.50                    & 0.005                    \\ \hline
\end{tabular}}}
\end{table}

\subsection{Results}
Once the predictions are obtained from the models, we applied the three detection rules based on different assumptions. The detection performance obtained on one of the time-sequences from the New York City data set is given in Figs. \ref{fig1} - \ref{fig3}. The numerical results obtained on evaluating the detection rules using Precision, Recall and F1-score are available in Table \ref{results}, along with the values of the false-positive regulators, $\tau_g$ and $q$. 

The false-positive regulators are the parameters that impact the performance of the detection algorithms. The false-positive regulator for the Gaussian-based detection rule, $\tau_g$, is chosen for each time-sequence such that the F1-score on the validation errors is maximized. The false-positive regulator for the EVT-based anomaly detection, $q$, is set from an initialization data stream. An initial threshold $t$ has to be chosen for the EVT-based detection, typically 98\% quantile. We set $q$ using the same initialization stream that is used for setting $t$.  The initialization stream contains the prediction errors from the training and validation sets. The probability $q$ is chosen so that the EVT-based anomaly detection picks up all the anomalies from the initialization stream. We observe that the threshold $\tau_{g}$ has a higher variability compared to that of the probability $q$. While $q$ remains in the range [$10^{-3}$, $10^{-5}$], $\tau_{g}$ varies between -15 and -25. Further, while a single $q$ value is sufficient for different time-sequences from the same data set (e.g., New York City Taxi Demand), different $\tau_{g}$ values are required for different streams of data from the same set. This translates into a relatively slower parameter tuning for the Gaussian-based detection on comparison with that of the EVT-based detection. 

Tukey's method is able to detect most of the anomalies but results in a large number of false-positives, which is not desirable. In other words, Tukey's method has a high Recall, but poor Precision. Only in the ECG data set, the Tukey's method achieves better prediction performance than the others. The fraction of anomalies is higher in the ECG data set, and hence, the anomalies cover a large spectrum above the upper quartile. Since the Tukey's method thresholds the raw prediction errors based on the upper quartile, it results in good anomaly detection for the ECG data set. 

Regardless of the application domain, we see that EVT-based detection rules provide consistently better performance that Gaussian-based and Tukey's method based detection rules. These findings suggest that presuming a Gaussian distribution on the prediction errors is a very strong assumption and might not hold for several scenarios. A more sensible assumption would be to assume that the tails follow GPD, which appears to be valid across diverse settings. On the other hand, assuming no distribution can result in multiple false alarms.

\section{Contributions} \label{conclusions}
Across application domains, accurate detection of abnormal patterns plays a vital role in the construction of reliable prediction algorithms. In this paper, we compared three detection rules that can be used with deep learning based anomaly detection, in the context of transportation networks. Each detection rule makes specific assumptions about the distribution of the prediction errors obtained from the Long Short-Term Memory (LSTM) network. Using statistical tests and numerical analysis, we showed that the widely used Gaussian distribution assumption on the prediction errors need not always hold. However, the tails distributions of the prediction errors are seen to follow a Generalized Pareto Distribution (GPD). This statistical evidence prompted us to devise a set of detection rules based on Extreme Value Theory (EVT).

The EVT-based detection rule consistently achieved more accurate anomaly detection compared to the Gaussian-based detection rule and Tukey's method. More variability was observed in the false-positive regulator values of the Gaussian-based detection rule compared to that of the EVT-based detection rule. The Gaussian-based detection required fixing of different false-positive regulator values for different sequences from the same data set, which in turn necessitated extensive parameter tuning. On the other hand, the EVT-based rule needed only a single value of false-positive regulator to achieve good performance across multiple streams from the same data set.

This paper follows an LSTM-based hybrid approach for anomaly detection. To get a comprehensive overview of various anomaly detection techniques, one should conduct an extensive comparison study of statistical, machine learning, and deep learning based models. Such a study is the next natural avenue for future research. Further, we aim to develop an end-to-end deep anomaly detection model by directly modifying the objective function of the LSTM to detect the anomalies.



\end{document}